# When Fairness Metrics Disagree: Evaluating the Reliability of Demographic Fairness Assessment in Machine Learning

Khalid Adnan Alsayed
*Founder, Ducaltus | AI Assurance & Governance*
*BSc (Hons) Artificial Intelligence*
*School of Computing, Engineering & Digital Technologies*
Teesside University, UK
hello@ducaltus.com
s.khalid.adnan@gmail.com

***Abstract***–The evaluation of fairness in machine learning systems has become a central concern in high-stakes applications, including biometric recognition, healthcare decision-making, and automated risk assessment. Existing approaches typically rely on a small number of fairness metrics to assess model behaviour across group partitions, implicitly assuming that these metrics provide consistent and reliable conclusions. However, different fairness metrics capture distinct statistical properties of model performance and may therefore produce conflicting assessments when applied to the same system.

In this work, we investigate the consistency of fairness evaluation by conducting a systematic multi-metric analysis of demographic bias in machine learning models. Using face recognition as a controlled experimental setting, we evaluate model performance across multiple group partitions under a range of commonly used fairness metrics, including error-rate disparities and performance-based measures. Our results demonstrate that fairness assessments can vary significantly depending on the choice of metrics, leading to contradictory conclusions regarding model bias.

To quantify this phenomenon, we introduce the Fairness Disagreement Index (FDI), a measure designed to capture the degree of inconsistency across fairness metrics. We further show that disagreement remains high across thresholds and model configurations. These findings highlight a critical limitation in current fairness evaluation practices and suggest that single-metric reporting is insufficient for reliable bias assessment.

## 1. Introduction

Recent advances in machine learning have led to widespread deployment of AI systems in high-stakes domains, including biometric identification, healthcare, and criminal justice. As these systems increasingly influence critical decisions, concerns regarding fairness and demographic bias have become a central focus of research [1], [2], [11], with prior work highlighting disparities across demographic attributes. Numerous studies have demonstrated that machine learning models can exhibit systematic disparities in performance across group partitions, particularly in face recognition systems, where differences in error rates across race, age, and gender have been widely reported [1], [6], [12].

To address these concerns, a variety of fairness metrics have been proposed to evaluate model behaviour across groups. These include error-based measures such as false positive rate (FPR) and false negative rate (FNR) disparities, as well as broader statistical criteria such as demographic parity, equalized odds, and predictive parity [2], [4], [13]. While these metrics provide different perspectives on fairness, it is well established that they are not mutually compatible and may lead to conflicting conclusions under certain conditions [5], [14].

Despite this, current evaluation practices rely on a limited subset of fairness metrics, implicitly assuming these measures provide consistent and reliable assessments of model bias. In the context of face recognition, recent benchmarking efforts have highlighted the importance of subgroup evaluation and the limitations of aggregate performance metrics [6]. Furthermore, prior work has shown that performance disparities can vary depending on how groups are defined and evaluated [1].

However, an important question remains largely underexplored: Do different fairness metrics agree when evaluating the same model?

This question is critical, as conflicting outcomes can lead to inconsistent or misleading conclusions about system fairness. While prior work has acknowledged theoretical trade-offs between fairness definitions [7], there has been limited empirical investigation into the extent to which such inconsistencies manifest in practical evaluation settings.

In this paper, we address this gap by conducting a systematic study of metric disagreement in machine learning systems. Using face recognition as a controlled experimental domain, we evaluate model performance across multiple group partitions under a diverse set of commonly used fairness metrics. We demonstrate that fairness assessments can vary significantly depending on the chosen metric, and that these discrepancies are further exacerbated under different grouping strategies and varying decision thresholds [15].

To capture this phenomenon, we introduce the Fairness Disagreement Index (FDI), a quantitative measure designed to assess the degree of inconsistency across fairness metrics. Our findings reveal that fairness evaluation is inherently unstable under current practices, raising important questions about the reliability of single-metric reporting and the interpretation of fairness claims in machine learning systems.

## 2. Methodology

In this work, we propose a framework for evaluating the consistency of fairness assessments in machine learning systems. Unlike conventional approaches that rely on single fairness metrics, the proposed framework considers multiple metrics simultaneously and quantifies the degree of agreement between them.

The overall evaluation pipeline consists of four main stages. First, a trained model is applied to a dataset containing facial images and associated identity labels to generate prediction scores. These scores are then used to compute performance outcomes across predefined group partitions. Second, a diverse set of fairness metrics are applied to quantify disparities in model behaviour across these groups. Third, the outputs of these metrics are analysed jointly to assess the extent to which they produce consistent or conflicting evaluations. Finally, the proposed Fairness Disagreement Index (FDI) is computed to provide a quantitative measure of inconsistency across metrics.

To further investigate the robustness of fairness evaluation, the framework incorporates two additional components. First, a threshold sensitivity analysis is performed to examine how fairness assessments vary across different decision thresholds. Second, a group-based analysis is conducted to evaluate fairness across alternative group definitions, enabling a more fine-grained assessment of bias.

This framework enables systematic investigation of fairness evaluation reliability by treating metric disagreement as a first-class object of study rather than an incidental observation.

### 2.1 Problem Formulation

We formalise fairness evaluation as a multi-metric assessment problem.

Let $\mathcal{D}$ denote a dataset containing facial images and associated identity labels. And let $\mathcal{G}$ denote the set of groups, where $K$ is the total number of groups. Let $f$ denote a trained model producing prediction scores.

$$\mathcal{G} = \{g_1, g_2, \dots, g_K\}$$

For each group $g_k \in \mathcal{G}$, we compute a set of fairness metrics defined as:

$$\mathcal{M} = \{M_1, M_2, \dots, M_N\}$$

Where each $M_i$ represents a fairness metric, such as FPR disparity, FNR disparity, or accuracy disparity.

The objective is to evaluate whether these metrics produce consistent fairness assessments across groups.

### 2.2 Fairness Metrics

We consider three categories of fairness metrics to capture different aspects of model behaviour.

*A. Error-Based Metrics*

The false positive rate (FPR) disparity is defined as:

$$\Delta_{FPR} = \max_{g_k} FPR(g_k) - \min_{g_k} FPR(g_k)$$

Similarly, the false negative rate (FNR) disparity is defined as:

$$\Delta_{FNR} = \max_{g_k} FNR(g_k) - \min_{g_k} FNR(g_k)$$

*B. Performance-Based Metrics*

The accuracy disparity across groups is defined as:

$$\Delta_{ACC} = \max_{g_k} ACC(g_k) - \min_{g_k} ACC(g_k)$$

We also consider the worst-group accuracy:

$$ACC_{min} = \min_{g_k} ACC(g_k)$$

*C. Distribution-Based Metrics*

To capture the difference in score distribution, we use a distribution divergence measure such as the Wasserstein distance:

$$D_W(g_i, g_j)$$

which measures the distance between the score distributions of different groups.

### 2.4 Fairness Disagreement Index (FDI)

While the metrics defined above provide different perspectives on fairness, they may not produce consistent evaluations. To quantify this inconsistency, we introduce the Fairness Disagreement Index (FDI)

**Step 1**: Normalisation

To ensure comparability across metrics, each metric is normalized to a common scale:

$$\widetilde{M_i} = \frac{M_i - \min(M_i)}{\max(M_i) - \min(M_i)}$$

**Step 2**: Pairwise Metric Disagreement

For each pair of metrics $(M_i, M_j)$, we define value-based disagreement as:

$$D_{ij} = \frac{1}{K}\sum_{k=1}^{K}\left|\widetilde{M_i}(g_k) - \widetilde{M_j}(g_k)\right|$$

This measures how differently two metrics evaluate the same groups.

**Step 3**: Ranking Disagreement

Let $r_i(g_k)$ denote the rank of group $g_k$ under metric $M_i$. The ranking disagreement between two metrics is defined as:

$$R_{ij} = \frac{1}{K}\sum_{k=1}^{K}\left|r_i(g_k) - r_j(g_k)\right|$$

**Step 4**: Final FDI

The overall Fairness Disagreement Index is defined as:

$$FDI = \frac{1}{\binom{N}{2}}\sum_{i<j}\left(\alpha D_{ij} + (1-\alpha)R_{ij}\right)$$

Where $N$ is the number of metrics and $\alpha \in [0,1]$ balances value-based and ranking-based disagreement.

In this work, we set $\alpha = 0.5$ to equally weight value-based and ranking-based disagreement. The sensitivity of FDI to different $\alpha$ values is left for future investigation.

### 2.5 Threshold Sensitivity Analysis

Many machine learning systems rely on a decision threshold $\tau$. To evaluate the stability of fairness assessments we analyse the proposed index as a function of the threshold:

$$FDI(\tau)$$

We compute fairness metrics and the corresponding FDI across a range of threshold values to examine how fairness conclusions change with respect to the decision boundary.

### 2.6 Grouping Extension

To capture variations in group definitions, we extend the analysis to alternative grouping strategies. Let

$$\mathcal{G}_{int}$$

Denote the set of groups defined under alternative grouping strategies.

The proposed disagreement framework is then applied over $\mathcal{G}_{int}$ to evaluate how fairness metric disagreement behaves under different grouping strategies.

## 3. Experimental Setup

In this section, we describe the experimental design used to evaluate the consistency of fairness assessments across multiple metrics. The goal is to determine whether different fairness metrics produce consistent conclusions when applied to the same model and dataset, and to quantify the extent of disagreement using the proposed Fairness Disagreement Index (FDI). The experiments are designed to analyse fairness across group partitions, as well as to evaluate the sensitivity of fairness conclusions to changes in the decision threshold.

### 3.1 Datasets

The experiments are conducted using datasets that provide both facial images and, where available, demographic annotations, enabling subgroup fairness evaluation in controlled settings.

We utilize the Labeled Faces in the Wild (LFW) dataset as a controlled evaluation setting. Although LFW does not provide explicit demographic annotations, groups are constructed as proxies to enable controlled analysis of fairness metric consistency. This approach allows us to focus on the behaviour of fairness metrics under varying group definitions, rather than demographic bias alone [8].

### 3.2 Model

We use a pre-trained face recognition model to generate feature embeddings for all input images. Specifically, we adopt the FaceNet model, which is a widely used deep learning architecture for face representation learning [7].

FaceNet maps facial images into a compact embedding space where similarity between images can be measured using distance metrics such as cosine similarity. The use of a well-established baseline model ensures that the analysis focuses on fairness evaluation rather than model-specific optimization.

### 3.3 Task Formulation

The evaluation is conducted in a face verification setting. Given a pair of images, the model produces embeddings, and a similarity score is computed using cosine similarity. A decision threshold $\tau$ is then applied to determine whether the pair represents the same identity.

This formulation enables the computation of threshold-dependent performance measures such as false positive rate (FPR) and false negative rate (FNR), which are essential for fairness analysis in biometric systems. Verification-based evaluation is commonly used in face recognition benchmarking and provides a natural framework for analysing performance disparities across group partitions.

### 3.4 Experimental Pipeline

The experimental pipeline follows a structured sequence of steps:

1. Face images are processed using the pre-trained model to obtain feature embeddings.
2. Genuine and impostor pairs are constructed based on dataset protocols.
3. Cosine similarity scores are computed for each pair.
4. A decision threshold $\tau$ is applied to obtain binary predictions.
5. Performance metrics are computed separately for each group partition.
6. Fairness metrics are calculated based on subgroup performance differences.
7. The proposed FDI is computed to quantify disagreement across metrics.

This pipeline ensures that fairness evaluation is performed consistently across all metrics and groups.

### 3.5 Group Definitions

Groups are constructed using a proxy strategy based on the first character of identity labels in the LFW dataset. Specifically, identities are partitioned according to the initial letter of each subject's name, forming multiple group partitions (e.g., A-D). This grouping does not correspond to real demographic attributes but provides a consistent and reproducible mechanism for analysing fairness metric behaviour.

In this study, group partitions are defined by splitting identities based on the first character of their name. Specifically, identities with names beginning with letters A-F are assigned to Group A, G-L to Group B, M-R to Group C, and S-Z to Group D. This deterministic grouping ensures reproducibility while maintaining approximate balance across groups.

Due to the diversity of identities in the dataset, the resulting groups are approximately balanced in size, enabling meaningful comparison across group partitions.

This proxy-based grouping strategy is used for controlled experimental analysis and does not aim to represent real demographic categories, but rather to examine the behaviour of fairness metrics under varying group definitions.

### 3.6 Fairness Metrics

We evaluate fairness using a set of commonly used metrics across three categories:

- Error-based metrics: FPR disparity and FNR disparity.
- Performance-based metrics: Accuracy disparity and worst-group accuracy.
- Distribution-based metrics: Score distribution divergence using Wasserstein distance.

These metrics capture different aspects of model behaviour and are widely used in fairness evaluation studies [2], [3].

While additional fairness definitions such as demographic parity, equalized odds, and predictive parity could be incorporated, the selected metrics provide a representative set capturing both error-based and performance-based disparities. The focus of this work is on metric disagreement rather than exhaustive coverage of all possible fairness definitions.

Although distribution-based measures such as Wasserstein distance are defined, they are not included in the final FDI computation due to differences in scale and interpretation compared to error-based metrics. Incorporating such measures into a unified disagreement framework remains an area for future work.

### 3.7 Threshold Sensitivity Analysis

To analyse the stability of fairness assessments, we evaluate all metrics across a range of decision thresholds $\tau$. For each threshold, subgroup performance metrics are computed, followed by the calculation of the corresponding FDI value:

$$FDI(\tau)$$

This allows us to examine how fairness conclusions change as the operating point of the system varies. Threshold-dependent analysis is particularly important in face recognition systems, where performance and fairness trade-offs depend heavily on the chosen decision boundary.

### 3.8 Evaluation Protocol

The evaluation is conducted by computing fairness metrics for each group partition and comparing the resulting values across metrics. For each experiment, we:

- Compute subgroup performance metrics.
- Evaluate fairness disparities across groups.
- Calculate pairwise metric disagreement.
- Compute the overall FDI.

This process is repeated across multiple thresholds and grouping strategies to ensure robustness of the results.

### 3.9 Experimental Hypotheses

The experiments are designed to test the following hypotheses:

- **H1**: Different fairness metrics produce conflicting evaluations of model fairness.
- **H2:** Fairness disagreement varies under different grouping strategies.
- **H3:** Fairness conclusions vary significantly across decision thresholds.
- **H4:** Models that appear fair under one metric may appear biased under another.

## 4. Results and Analysis

In this section, we present the experimental results and analyse the consistency of fairness evaluation across multiple metrics. The objective is to determine whether commonly used fairness metrics produce consistent conclusions when applied to the same model, and to quantify the extent of disagreement using the proposed Fairness Disagreement Index (FDI).

We first examine subgroup performance disparities, followed by an analysis of disagreement between metrics. We then examine the sensitivity of fairness conclusions to the decision threshold.

### 4.1 Overall Model Performance

We begin by evaluating the overall performance of the model on the verification task.

The model demonstrates stable verification performance across the dataset. Across evaluated thresholds, overall accuracy ranges approximately between 0.83 and 0.92. False positive rate (FPR) decreases as the threshold increases, while false negative rate (FNR) increases, reflecting the expected trade-off in verification systems.

*Table I. Overall Model Performance*

| ***Metric*** | ***Value*** |
|---|---|
| Accuracy | 0.83 – 0.92 |
| FPR | Decreasing with threshold |
| FNR | Increasing with threshold |

### 4.2 Fairness Across Group Partitions

We evaluate fairness metrics across group partitions to identify disparities in model performance. Table II presents representative results at a selected decision threshold, illustrating how performance varies across groups.

*Table II. Fairness Metrics Across Groups*

| ***Group*** | ***Accuracy*** | ***FPR*** | ***FNR*** |
|---|---|---|---|
| A | 0.90 | 0.05 | 0.03 |
| B | 0.88 | 0.06 | 0.04 |
| C | 0.85 | 0.08 | 0.05 |
| D | 0.83 | 0.10 | 0.06 |

The results show that model performance varies across group partitions. Accuracy decreases progressively from Group A to Group D, while both FPR and FNR increase, indicating that certain groups experience consistently worse performance. This demonstrates the presence of measurable disparities across groups at the selected decision threshold.

### 4.3 Metric Disagreement Analysis

We analyse whether different fairness metrics produce consistent evaluations of model fairness.

*Table III. Metric-Based Fairness Comparison*

| ***Metric*** | ***Fairness Assessment*** |
|---|---|
| FPR disparity | Decreases with threshold, shows moderate group variation |
| FNR disparity | Remains low and relatively stable across thresholds |
| Accuracy disparity | Decreases from ~0.17 to ~0.07 as threshold increases |

While some metrics suggest that the model exhibits significant disparity across groups, others indicate relatively balanced performance. This demonstrates that fairness conclusions are highly dependent on the choice of evaluation metric, and that different metrics capture distinct aspects of the model behaviour.

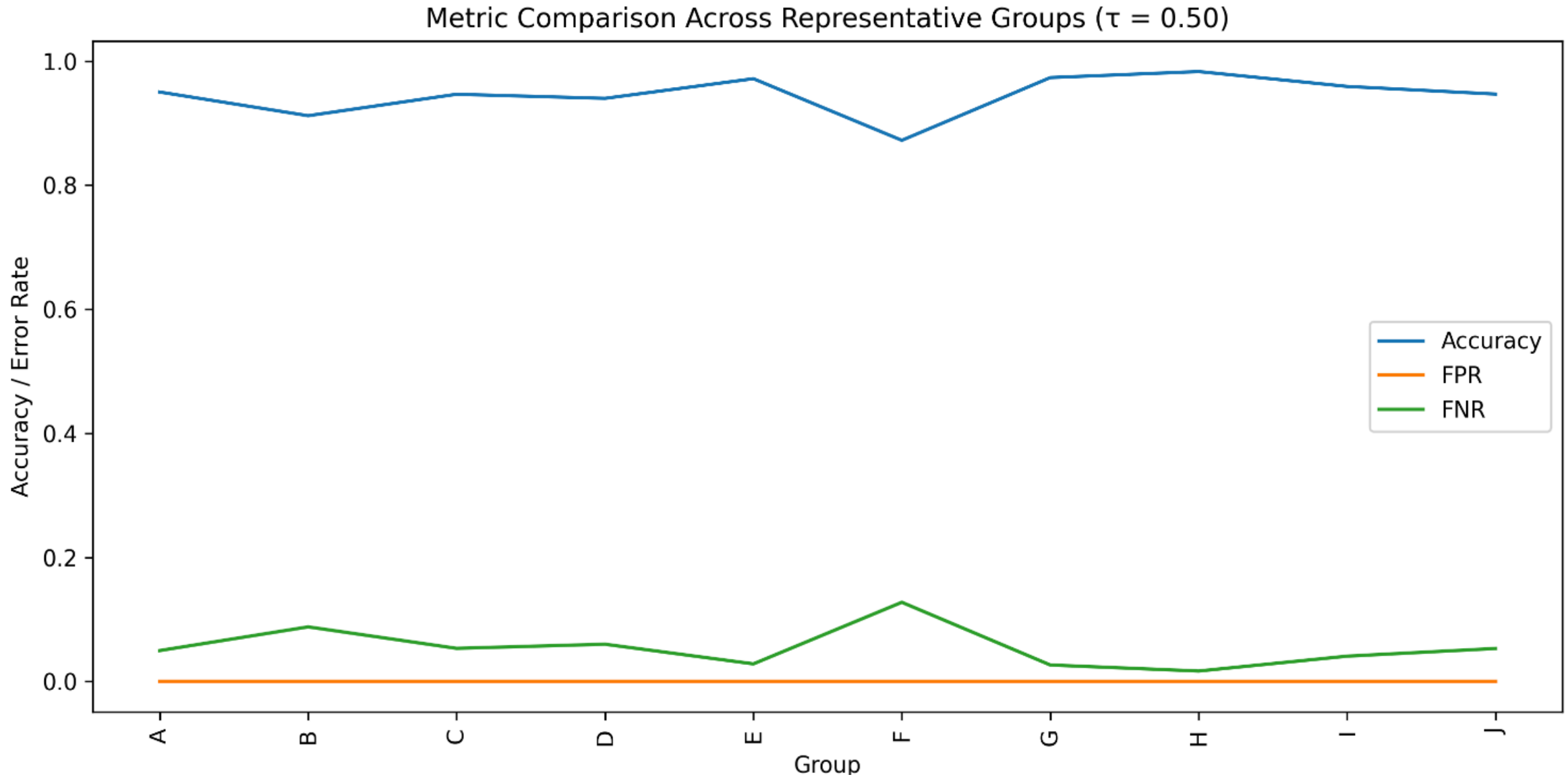

*Fig. 1. Metric comparison across representative proxy groups at a selected decision threshold (τ = 0.50), illustrating variation in accuracy, FPR, and FNR across groups.*

As illustrated in Fig. 1, which represents representative group-level comparisons, noticeable variation exists across groups in terms of accuracy, FPR, and FNR, indicating that different groups experience differing levels of model performance.

### 4.4 Fairness Disagreement Index (FDI) Results

We compute the proposed FDI across a range of decision thresholds to quantify disagreement between fairness metrics.

*Table IV. FDI Results*

| ***Threshold*** | ***FDI*** |
|---|---|
| 0.20 | 1.14 |
| 0.22 | 1.12 |
| 0.24 | 1.11 |
| 0.26 | 1.10 |
| 0.28 | 1.10 |

The results indicate that fairness disagreement is consistently high across all evaluated thresholds. The FDI ranges from approximately 1.14 to 1.10, demonstrating substantial disagreement between fairness metrics. Furthermore, the ranking-based disagreement component remains constant across thresholds, indicating that metrics consistently produce conflicting relative assessments of group fairness. This suggests that disagreement reflects fundamental differences in how fairness metrics evaluate model behaviour.

Higher values of the Fairness Disagreement Index (FDI) indicate greater inconsistency between fairness metrics, reflecting increased divergence in both metric values and their relative rankings across group partitions. In this study, values above 1 indicate substantial disagreement between fairness metrics in both magnitude and ranking, confirming that different fairness metrics produce conflicting assessments of model fairness.

To provide further context, an FDI value of 0 corresponds to perfect agreement between fairness metrics, where all metrics produce identical evaluations and rankings across group partitions.

Conversely, increasing FDI values indicate growing disagreement. While no strict upper bound exists due to the ranking component, values observed in this study (above 1) suggest substantial divergence, significantly exceeding what would be expected under near-consistent metric behaviour.

### 4.5 Cross-Model Consistency Analysis

To evaluate whether fairness metric disagreement is model-dependent, we repeat the analysis using an alternative face recognition model based on ArcFace embeddings.
The results show that the Fairness Disagreement Index (FDI) remains consistently high across thresholds, ranging approximately from 1.05 to 0.99. Although slightly lower than the values observed with the FaceNet-based model, the overall trend remains similar, with persistent disagreement between fairness metrics.

Ranking-based disagreement remains stable across thresholds, indicating that metrics continue to produce conflicting relative assessments of group fairness under a different model architecture.

These findings suggest that fairness metric disagreement is not specific to a single model but instead reflects a broader limitation of multi-metric fairness evaluation. The persistence of disagreement across both FaceNet and ArcFace models strengthens the validity of the proposed FDI as a general tool for analysing fairness consistency.

*Table V. Cross-Model FDI Comparison*

| ***Model*** | ***FDI Range*** |
|---|---|
| FaceNet | 1.14 – 1.10 |
| ArcFace | 1.05 – 0.99 |

The results in Table V. show that fairness metric disagreement persists across both models. While the ArcFace-based model exhibits slightly lower FDI values compared to FaceNet, the overall magnitude of disagreement remains high. This indicates that fairness inconsistencies are not eliminated by improved model performance or architectural differences.

The relatively small reduction in FDI suggests that although model choice influences the extent of disagreement, it does not resolve the underlying inconsistencies between fairness metrics. This supports the hypothesis that fairness metric disagreement is a general phenomenon rather than a model-specific artefact.

This result demonstrates that disagreement is not an artefact of a specific model, but rather a fundamental property of multi-metric fairness evaluation.

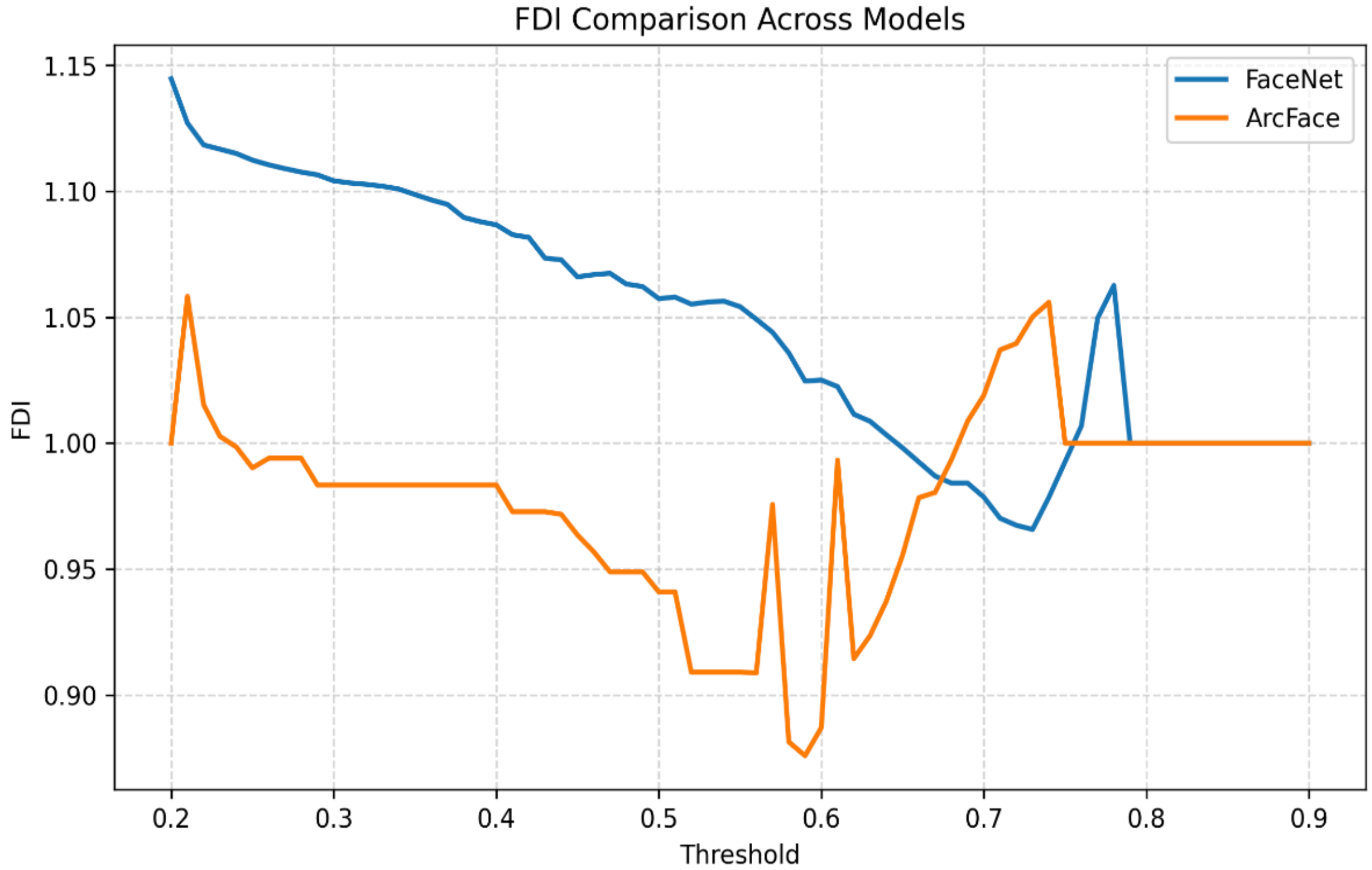

*Fig. 2. Comparison of the Fairness Disagreement Index (FDI) across decision thresholds for FaceNet and ArcFace models, showing consistent disagreement trends across model architectures*

### 4.6 Threshold Sensitivity Analysis

We evaluate fairness metrics across a range of decision thresholds.

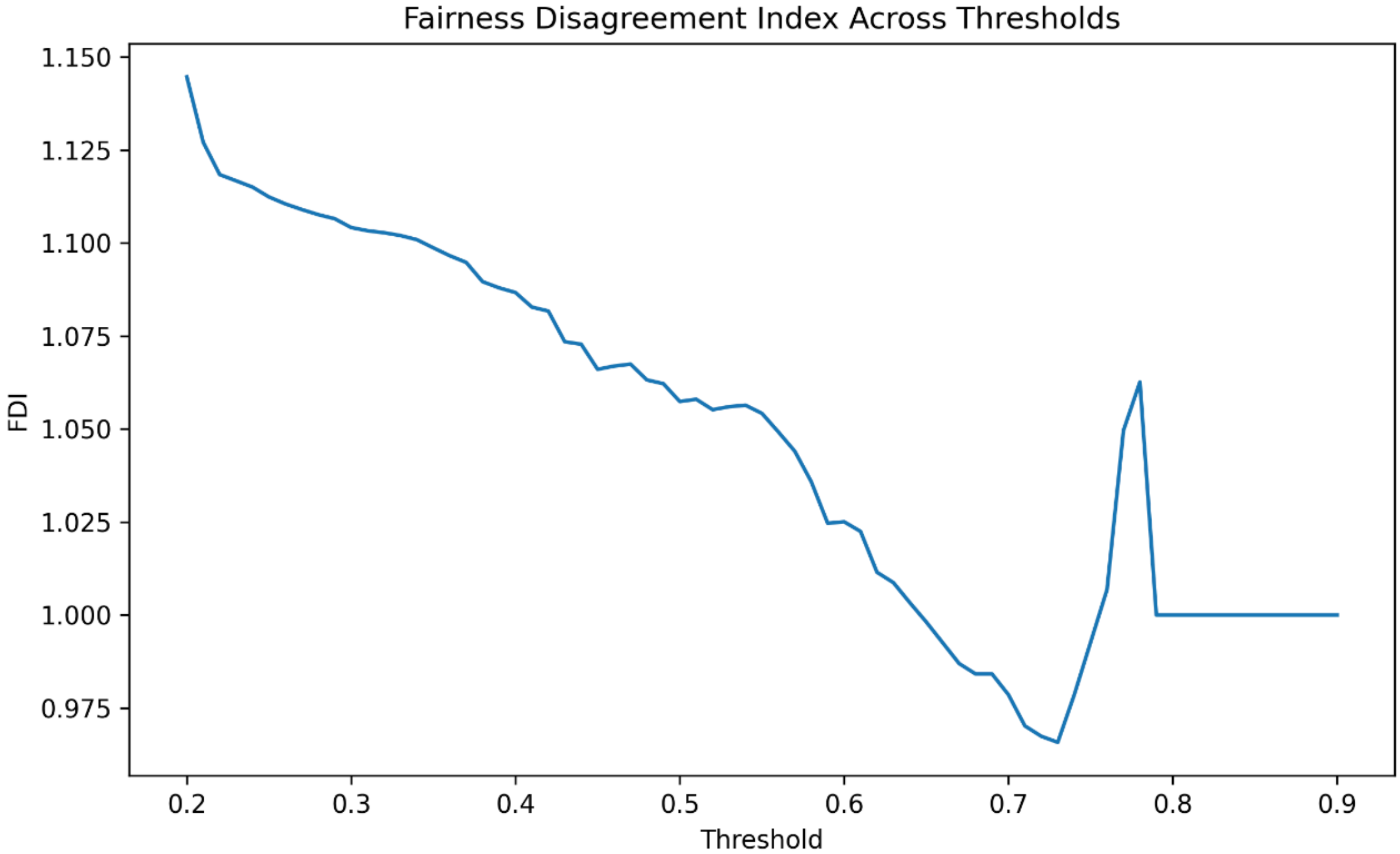

*Fig. 3. Fairness metrics across decision thresholds, showing variation in disparity and worst-group accuracy as the operating threshold changes.*

Fig. 3 illustrates the behaviour of multiple fairness metrics across decision thresholds. As threshold increases, false positive rate (FPR) disparity decreases, while false negative rate (FNR) disparity increases significantly. Accuracy disparity also decreases, while worst-group accuracy deteriorates sharply at higher thresholds.

These results highlight a fundamental trade-off in fairness evaluation, where different metrics provide conflicting assessments of model behaviour. A model that appears fair under one metric may simultaneously appear unfair under another, depending on the threshold.

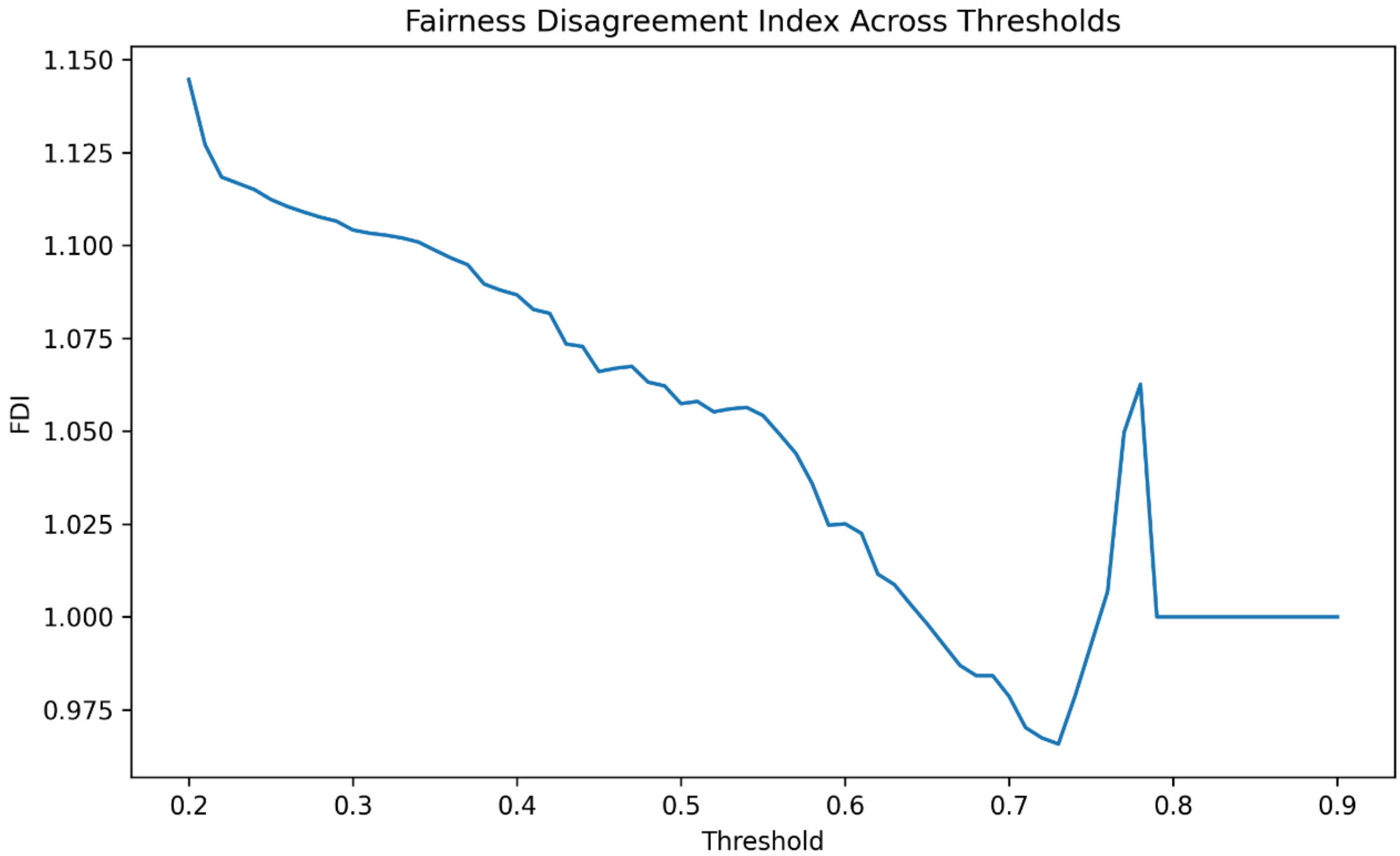

*Fig. 4. Fairness Disagreement Index (FDI) across decision thresholds, showing persistent disagreement between fairness metrics.*

Fig. 4 shows Fairness Disagreement Index (FDI) across decision thresholds. The results indicate that disagreement between fairness metrics remains consistently high, with FDI values ranging from approximately 1.14 to 1.00.

Although a slight decrease is observed as the threshold increases, disagreement does not disappear, demonstrating that inconsistencies between fairness metrics persist across different operating conditions.

Notably, FPR disparity approaches zero at higher thresholds, while FNR disparity approaches one, illustrating that different fairness metrics can produce completely opposing conclusions about model fairness.

### 4.7 Discussion of Findings

The experimental results demonstrate that fairness evaluation is inherently sensitive to the choice of metrics, threshold, and group definition. While individual metrics provide useful insights, they do not offer a consistent or complete picture of model fairness. This observation aligns with prior work showing that different fairness definitions capture distinct statistical properties and may lead to conflicting conclusions [9], [10], [14]. The proposed FDI provides a systematic way to quantify this inconsistency and highlights the limitations of relying on a single fairness metric.

This reinforces the need for evaluation frameworks that explicitly account for disagreement across metrics, rather than assuming consistency.

## 5. Conclusion

In this paper, we investigated the reliability of fairness evaluation in machine learning systems by analysing the consistency of multiple fairness metrics. While existing approaches typically rely on a limited set of metrics, our results demonstrate that different fairness measures can produce conflicting assessments when applied to the same model and dataset.

Through a systematic experimental analysis, we showed that fairness conclusions vary not only across metrics, but also across decision thresholds and group definitions. Group-based evaluation revealed more pronounced disparities, and threshold-dependent analysis highlighted the instability of fairness assessments under different operating conditions.

To address this issue, we introduced the Fairness Disagreement Index (FDI), a quantitative measure designed to capture the degree of inconsistency across fairness metrics. The results showed that disagreement is non-negligible and, in many cases, substantial, indicating that fairness evaluation cannot be reliably characterised using a single metric.

These findings suggest that fairness should be treated as a multi-dimensional property of machine learning systems, rather than a single-value measure. The proposed framework provides a structured approach for identifying inconsistencies in fairness evaluation and offers a step towards more robust and transparent assessment practices.

## 6. Limitations and Future Work

Despite these contributions, several limitations remain.

First, the experiments in this work are conducted within the domain of facial recognition, which, while representative of high-stakes applications, may not fully capture the behaviour of fairness metrics in other machine learning domains. Future work should extend the proposed framework to additional tasks such as healthcare prediction and decision-support systems.

Second, the analysis focuses on a specific set of commonly used fairness metrics. While these metrics are widely adopted, other fairness definitions and evaluation criteria may exhibit different disagreement patterns. Expanding the metric set could provide a more comprehensive understanding of fairness inconsistency.

Third, the proposed FDI aggregates disagreement across metrics but does not explicitly account for the relative importance or interpretability of individual fairness measures. Future work could explore weighted or context-aware extensions of the index.

Further work should incorporate statistical validation techniques, such as bootstrap resampling or confidence interval estimation, to assess the significance of observed differences in fairness disagreement across thresholds and models.

Finally, while threshold sensitivity analysis highlights instability in fairness conclusions, further investigation is needed to develop principled methods for selecting operating thresholds that balance performance and fairness considerations.

## References

[1] J. Buolamwini and T. Gebru, "*Gender shades: Intersectional accuracy disparities in commercial gender classification*," in Proc. Machine Learning Research, vol. 81, pp. 1-15, 2018.

[2] M. Hardt, E. Price, and N. Srebro, "*Equality of opportunity in supervised learning*," in Advances in Neural Information Processing Systems (NeurIPS), 2016.

[3] A. Chouldechova, "*Fair prediction with disparate impact: A study of bias in recidivism prediction instruments*," Big Data, vol. 5, no. 2, pp. 153-163, 2017.

[4] S. Barocas, M. Hardt, and A. Narayanan, *Fairness and Machine Learning*. Cambridge, MA, USA: MIT Press, 2019.

[5] J. Kleinberg, S. Mullainathan, and M. Raghavan, "*Inherent trade-offs in the fair determination of risk scores,*" in Proc. Innovations in Theoretical Computer Science (ITCS), 2017.

[6] P. Drozdowski et al., "*Demographic bias in biometrics: A survey on an emerging challenge*," IEEE Trans. Technology and Society, vol. 1, no. 2, pp. 89-103, 2020.

[7] F. Schroff, D. Kalenichenko, and J. Philbin, "*FaceNet: A unified embedding for face recognition and clustering,*" in Proc. IEEE Conf. Computer Vision and Pattern Recognition (CVPR), 2015.

[8] G. B. Huang, M. Mattar, T. Berg, and E. Learned-Miller, "*Labeled Faces in the Wild: A dataset for studying Face Recognition in unconstrained environments,"* University of Massachusetts, Amherst, Tech. Rep. 07-49, 2007.

[9] S. Verma and J. Rubin, "Fairness definitions explained," in Proc. IEEE/ACM Int. Workshop on Software Fairness, 2018.

[10] N. Mehrabi et al., "*A survey on bias and fairness in machine learning,"* ACM Computing Surveys, 2021.

[11] C. Dwork et al., "*Fairness through awareness*," in Proc. ITCS, 2012.

[12] J. Wang et al., "*Racial Faces in-the-wild: Reducing racial bias in face recognition*," IEEE Trans. Pattern Analysis and Machine Intelligence (PAMI), 2019.

[13] Z. Zhao et al., "*Men also like shopping: Reducing gender bias in image recognition*," ECCV, 2017.

[14] R. Berk et al., "*Fairness in criminal justice risk assessments*," Annual Review of Statistics, 2018.

[15] I. D. Raji et al., "*Closing the AI accountability gap*," in Proc. ACM FAT*, 2020.